\documentclass{article}

\usepackage{PRIMEarxiv}

\usepackage[utf8]{inputenc} 
\usepackage[T1]{fontenc}    
\usepackage{hyperref}       
\usepackage{url}            
\usepackage{booktabs}       
\usepackage{amsfonts}       
\usepackage{nicefrac}       
\usepackage{microtype}      
\usepackage{lipsum}
\usepackage{fancyhdr}       
\usepackage{graphicx}       
\graphicspath{{media/}}     

\usepackage{url,hyperref,lineno,microtype,subcaption}
\usepackage{multirow}
\usepackage{amsmath}

\usepackage{tabularx}
\usepackage{diagbox}

\title{A Human-Annotated Video Dataset for Training and Evaluation of 360-Degree Video Summarization Methods
\thanks{\textit{\underline{Citation}}: 
\textbf{I. Kontostathis, E. Apostolidis, V. Mezaris. An Integrated Framework for Multi-Granular Explanation of Video Summarization. Proc. 1st Int. Workshop on Video for Immersive Experiences (Video4IMX-2024) at ACM IMX 2024, Stockholm, Sweden, June 2024.}}
\thanks{© I. Kontostathis, E. Apostolidis, V. Mezaris, 2024. This is the author's version of the work. It is posted here for
your personal use. Not for redistribution. The definitive version was published in the ACM IMX 2024 Workshop Proceedings.}

}

\author{
  Ioannis Kontostathis, Evlampios Apostolidis, Vasileios Mezaris \\
  Information Technologies Institute (ITI), Centre for Research and Technology, Hellas (CERTH) \\
  Thessaloniki, Greece\\
  \texttt{\{ioankont, apostolid, bmezaris\}@iti.gr} \\
}

\begin{document}
\maketitle

\begin{abstract}
In this paper we introduce a new dataset for 360-degree video summarization: the transformation of 360-degree video content to concise 2D-video summaries that can be consumed via traditional devices, such as TV sets and smartphones. The dataset includes ground-truth human-generated summaries, that can be used for training and objectively evaluating 360-degree video summarization methods. Using this dataset, we train and assess two state-of-the-art summarization methods that were originally proposed for 2D-video summarization, to serve as a baseline for future comparisons with summarization methods that are specifically tailored to 360-degree video. Finally, we present an interactive tool that was developed to facilitate the data annotation process and can assist other annotation activities that rely on video fragment selection.
\end{abstract}


\section{Introduction}

Nowadays, there is an increasing interest in the production and distribution of $360^{\circ}$ video content, in order to offer a more comprehensive and immersive viewing experience to the users. This interest is supported by the existence of advanced $360^{\circ}$ video recording devices (e.g., GoPro and Insta360) and the compatibility of the most popular social networks and video sharing platforms with this type of video content. Currently, the consumption of $360^{\circ}$ videos is being made primarily using VR headsets (e.g., Oculus Rift and Gear VR). Nevertheless, transforming $360^{\circ}$ videos to concise 2D-video summaries that can be viewed via traditional devices, such as TV sets and smartphones, would: i) enable repurposing, ii) increase consumption (through additional devices), iii) and facilitate browsing and retrieval of $360^{\circ}$ video content. This observation indicates a strong need for technologies that could support the summarization of $360^{\circ}$ videos, and highlights the importance of these technologies in the modern video content production and consumption ecosystem. 

As reported in \cite{kontostathis2024summarization}, currently, the research focuses mainly on methods for navigating the viewer through the unlimited field of view of the $360^{\circ}$ video, by controlling the camera's position and field of view, and generating an optimal camera trajectory \cite{su2016activity,Su_2017_CVPR,Hu_2017_CVPR,9072511,10.1145/3306346.3323046,9284734}. The output is a normal Field-Of-View (NFOV) video that has the same duration with the $360^{\circ}$ video, and thus cannot be seen as a summary of the video content. Only a few recent works deal with the summarization of $360^{\circ}$ videos \cite{Yu2019DeepRanking,Lee_2018_CVPR,kontostathis2024summarization}. Nevertheless, the methods in \cite{Yu2019DeepRanking,Lee_2018_CVPR} assume the existence of a single important activity or narrative in the $360^{\circ}$ video and target the creation of a video highlight \cite{Yu2019DeepRanking} or a story-based video summary \cite{Lee_2018_CVPR}, respectively. In terms of data, the majority of the used datasets in these works \cite{su2016activity,Hu_2017_CVPR,8418756,Xu_2018_CVPR} can assist the training of network architectures for NFOV selection and saliency prediction, and thus can support the development of methods for $360^{\circ}$ video summarization, only partially. Moreover, the used datasets in works for $360^{\circ}$ video highlight detection \cite{Yu2019DeepRanking} and story-based summarization \cite{Lee_2018_CVPR} (which are more closely related with the summarization task) are not publicly-available. Finally, due to lack of ground-truth data for $360^{\circ}$ video summarization, the performance of the method from \cite{kontostathis2024summarization} was evaluated only based on qualitative analysis.

To assist research on $360^{\circ}$ video summarization, in this paper we introduce a new dataset, called 360-VSumm, that
contains diverse visual content from different topics (music shows, sports games, short movies, documentaries) captured under various conditions (indoor, outdoor) using a static or moving camera. 360-VSumm contains ground-truth annotations about the parts of the $360^{\circ}$ video that should be included in the summary (differently from \cite{su2016activity,Hu_2017_CVPR,8418756,Xu_2018_CVPR}). Moreover, the included videos might contain multiple events that overlap in time or run in parallel; thus, the created 360-VSumm dataset is not tailored to specific summarization scenarios (contrary to \cite{Yu2019DeepRanking,Lee_2018_CVPR}) and can assist the development of generic $360^{\circ}$ video summarization methods. Our main contributions are as follows:
\begin{itemize}
    \item We introduce the 360-VSumm dataset for $360^{\circ}$ video summarization. This dataset includes ground-truth human-generated summaries, which can be used for training and objectively evaluating $360^{\circ}$ video summarization methods.
    \item We use 360-VSumm to train and evaluate two state-of-the-art summarization methods that were originally proposed for 2D-video summarization, to serve as a baseline for future comparisons with summarization methods that are specifically tailored to $360^{\circ}$ video.
    \item We develop an interactive tool for annotating videos by selecting their most appropriate parts for inclusion in the video summary, which can be used for annotation tasks that rely on video fragment selection (e.g., to create ground-truth data for video highlight detection and video abnormal event detection).
\end{itemize}

\section{Literature Review}
\label{sec:literature}

As discussed in \cite{kontostathis2024summarization}, the task of creating a natural-looking NFOV video that summarizes the interesting events of a $360^{\circ}$ video, has been studied only to a small extent using datasets that are tailored to the needs of specific NFOV or summary video production scenarios. Su et al. developed methods for extracting a singular NFOV video from a panoramic $360^{\circ}$ video \cite{su2016activity,Su_2017_CVPR}. In both of these works, Su et al. used the Pano2Vid dataset that was created by collecting $86$ $360^{\circ}$ videos from YouTube using specific keywords such as ``Hiking'', ``Mountain Climbing'', ``Parade'' and ``Soccer''. A subset of these videos ($20$ in total) was annotated by three different annotators with human-edited NFOV camera trajectories (two per video), and the obtained annotations were used as ground-truth for evaluating the performance of the developed methods. Hu et al. \cite{Hu_2017_CVPR} proposed a method for piloting the viewer in the $360^{\circ}$ video by detecting its most interesting parts. To evaluate their method, Hu et al. collected a new dataset, called Sports-360, by downloading from YouTube $342$ $360^{\circ}$ videos showing five sports activities (namely, basketball, parkour, BMX, skateboarding, and dance), and asking three annotators to label the most salient object for VR viewers in frames containing human-identifiable objects. Qiao et al. \cite{9072511}, presented a method for predicting viewport-dependent saliency over $360^{\circ}$ videos based on both video content and viewport location. In terms of training and evaluation, they used the PVS-HM dataset from \cite{8418756} and the VR-EyeTracking dataset from \cite{Xu_2018_CVPR}. The former contains $76$ panoramic video sequences of diverse content (e.g., driving, sports, movies, video games) along with data about the head movement and the eye fixation of $58$ humans. The latter is composed of $208$ $360^{\circ}$ videos of various content (e.g., indoor scene, outdoor activities, music shows, sports games, documentation, short movies) that were annotated with the help of $30$ humans in terms of eye fixation. Yu et al. \cite{Yu2019DeepRanking} proposed a framework for $360^{\circ}$ video highlight detection, that finds out the best NFOV subshot per $360^{\circ}$ video segment and selects the N top-ranked NFOV subshots as highlights. The evaluation of this framework was based on the Pano2vid dataset and a new $360^{\circ}$ video highlight dataset, which was formed after collecting $2406$ NFOV and $115$ $360^{\circ}$ wedding and music videos from YouTube and Vimeo. As ground-truth data, Yu et al. used the annotations ($15$ most salient 5-sec. NFOV video clips) of three annotators on $25$ randomly-selected $360^{\circ}$ videos from each category. Nevertheless, the created dataset was not released in public. Lee et al. \cite{Lee_2018_CVPR} described an approach for story-based summarization of $360^{\circ}$ videos. This approach identifies NFOV region proposals using a deep ranking network and performs temporal summarization based on a memory network and the assumption that the parts of a story-based summary share a common story-line. The NFOV selection part was evaluated using Pano2Vid dataset, while the temporal summarization part was assessed using a newly collected $360^{\circ}$ video dataset. This dataset comprises of $285$ $360^{\circ}$ videos and $9138$ photostreams showing five different topics (namely ``wedding'', ``parade'', ``rowing'', ``scuba diving'' and ``air ballooning''). These data are used in an unsupervised training setting in order to learn transitions between the subshots of a summary, so as to recover the latent storyline and summarize the full-length video. For evaluation, Lee et al. randomly sampled $10$ $360^{\circ}$ videos per topic as a test set and obtained three ground-truth summaries per video from human annotators, given that the generated summary does not exceed the $15\%$ of the full-length video's duration. However, the used dataset was not made publicly-available. Kang et al. \cite{10.1145/3306346.3323046} presented an interactive $360^{\circ}$ video navigation system, which defines a virtual camera path based on the most salient events in the video and generates a NFOV video. The performance of this system was evaluated with the help of the Pano2Vid \cite{su2016activity} and Sports-360 \cite{Hu_2017_CVPR} datasets. Focusing on a similar task, Wang et al. \cite{9284734} developed a tool for $360^{\circ}$ video navigation and playback on 2D displays, and assessed its performance using the same datasets. Finally, in our previous work \cite{kontostathis2024summarization}, we proposed an integrated system for spatio-temporal summarization of $360^{\circ}$ videos. This system detects salient events in the $360^{\circ}$ video using two state-of-the-art methods from the literature (ATSal \cite{Atsal_2020} and SST-Sal \cite{Sstsal_2022_CG}), creates a conventional 2D-video that contains these events, and produces a summary of this video using a saliency-aware variant of an unsupervised state-of-the-art video summarization method (CA-SUM \cite{10.1145/3512527.3531404}). For the training and evaluation of the $360^{\circ}$ video saliency detection component, we used the VR-EyeTracking and Sports-360 datasets. The training of the video summarization component was based on a set of $100$ 2D-videos that were produced and scored in terms of frame-level saliency using the integrated mechanism for salient event detection and 2D-video production, while its assessment was based on qualitative analysis. 

The aforementioned details about the used datasets in the relevant literature are summarized in Table \ref{tab:datasets}. As can be seen, most of these datasets \cite{su2016activity,Hu_2017_CVPR,8418756,Xu_2018_CVPR} can be used to train network architectures for NFOV selection and saliency prediction, in order to support the creation of NFOV videos from $360^{\circ}$ videos \cite{su2016activity,Su_2017_CVPR}, the navigation of the viewer in the content of $360^{\circ}$ videos \cite{Hu_2017_CVPR, 10.1145/3306346.3323046, 9284734}, and the prediction of viewport-dependent $360^{\circ}$ video saliency \cite{9072511}. These datasets can assist only partially the development of methods for $360^{\circ}$ video highlight detection \cite{Yu2019DeepRanking} or summarization \cite{Lee_2018_CVPR,kontostathis2024summarization}, since they can be used only for specific parts of the processing pipeline, that relate to NFOV selection or saliency prediction. Only a couple of datasets are suitable for training networks that create a more concise version of the content of the full-length $360^{\circ}$ video \cite{Yu2019DeepRanking,Lee_2018_CVPR}. However, both of them were formed based on the assumption that there is only one important activity or narrative in the $360^{\circ}$ video, and thus can be used only for building methods for $360^{\circ}$ video highlight detection \cite{Yu2019DeepRanking} or story-based summarization \cite{Lee_2018_CVPR}. Moreover, none of these datasets is publicly-available.

In this paper we introduce a new dataset, called 360-VSumm, that can assist research on $360^{\circ}$ video summarization. 360-VSumm is based on the VR-EyeTracking dataset \cite{Xu_2018_CVPR} and thus it comprises of diverse visual content from different topics (music shows, sports games, short movies, documentaries) that has been captured under various conditions (indoor, outdoor) using a static or moving camera. Differently from \cite{su2016activity,Hu_2017_CVPR,8418756,Xu_2018_CVPR}, our dataset contains ground-truth annotations about the parts of the $360^{\circ}$ video that should be included in the summary. Moreover, contrary to \cite{Yu2019DeepRanking,Lee_2018_CVPR}, the videos in our dataset might contain multiple events that overlap in time or run in parallel and should be all taken into account during the summarization process. So, 360-VSumm is not restricted to specific rules about the content of the $360^{\circ}$ video, and thus can assist the development of more generic $360^{\circ}$ video summarization methods. 

\begin{table}[t]
\centering
\caption{Overview of the used datasets in the relevant literature.}
\resizebox{\columnwidth}{!}{%
\begin{tabular}{|l|l|l|l|l|l|}
\hline
\textbf{Dataset}                                                                      & \textbf{\begin{tabular}[c]{@{}l@{}}Number and\\ type of videos\end{tabular}}   & \textbf{Visual content}                                                                                                                               & \textbf{Ground-truth annotations}                                                                                                            & \textbf{Supported task(s)}                                                                                           & \textbf{Availability}   \\ \hline
\multirow{4}{*}{Pano2Vid \cite{su2016activity}}                                                             & \multirow{4}{*}{$86$ $360^{\circ}$ videos}                                                 & \multirow{4}{*}{\begin{tabular}[c]{@{}l@{}}Hiking, Parade, \\Soccer, Mountain \\Climbing\end{tabular}}                                                  & \multirow{4}{*}{\begin{tabular}[c]{@{}l@{}}Human-edited NFOV\\ camera trajectories\end{tabular}}                                             & \begin{tabular}[c]{@{}l@{}}- NFOV video creation\\ from $360^{\circ}$ videos \cite{su2016activity,Su_2017_CVPR}\end{tabular}                                      & \multirow{4}{*}{Public} \\
                                                                                      &                                                                                &                                                                                                                                                       &                                                                                                                                              & \begin{tabular}[c]{@{}l@{}}- $360^{\circ}$ video highlight detection \cite{Yu2019DeepRanking}\\ (used to evaluate NFOV selection)\end{tabular}          &                         \\
                                                                                      &                                                                                &                                                                                                                                                       &                                                                                                                                              & \begin{tabular}[c]{@{}l@{}}- Story-based $360^{\circ}$ video summarization \cite{Lee_2018_CVPR}\\ (used to evaluate NFOV selection)\end{tabular}   &                         \\
                                                                                      &                                                                                &                                                                                                                                                       &                                                                                                                                              & - $360^{\circ}$ video viewer navigation \cite{10.1145/3306346.3323046, 9284734}                                                                                       &                         \\ \hline
\multirow{3}{*}{Sports-360 \cite{Hu_2017_CVPR}}                                                           & \multirow{3}{*}{$342$ $360^{\circ}$ videos}                                                & \multirow{4}{*}{\begin{tabular}[c]{@{}l@{}}Basketball, BMX, \\Dance, Parkour,\\Skateboarding\end{tabular}}                                             & \multirow{3}{*}{\begin{tabular}[c]{@{}l@{}}Labels about the most \\salient object in frames \\with human-identifiable \\objects\end{tabular}} & \multirow{2}{*}{- $360^{\circ}$ video viewer navigation \cite{Hu_2017_CVPR, 10.1145/3306346.3323046, 9284734}}                                                                       & \multirow{3}{*}{Public} \\
                                                                                      &                                                                                &                                                                                                                                                       &                                                                                                                                              &                                                                                                                      &                         \\
                                                                                      &                                                                                &                                                                                                                                                       &                                                                                                                                              & \begin{tabular}[c]{@{}l@{}}- $360^{\circ}$ video summarization \cite{kontostathis2024summarization} (used to\\ evaluate $360^{\circ}$ video saliency prediction)\end{tabular} &                         \\ \hline
PVS-HM \cite{8418756}                                                                               & \begin{tabular}[c]{@{}l@{}}$76$ panoramic\\ videos\end{tabular}                  & \begin{tabular}[c]{@{}l@{}}Driving, Sports,\\ Movies, Video games\end{tabular}                                                                        & \begin{tabular}[c]{@{}l@{}}Head movement and eye\\ fixation data for $58$ humans\end{tabular}                                                & \begin{tabular}[c]{@{}l@{}}Viewport-dependent $360^{\circ}$ video\\ saliency prediction \cite{9072511}\end{tabular}                           & Public                  \\ \hline
\multirow{1}{*}{VR-EyeTracking \cite{Xu_2018_CVPR}}      & \multirow{1}{*}{$208$ $360^{\circ}$ videos}  & \begin{tabular}[c]{@{}l@{}}Indoor scene, \\ Outdoor activities,\\ Music shows, \\ Sports games,\\ Short movies, \\ Documentation\end{tabular} & \multirow{1}{*}{\begin{tabular}[c]{@{}l@{}}Eye fixation data for $30$ \\humans\end{tabular}} 
     & \begin{tabular}[c]{@{}l@{}}- Viewport-dependent $360^{\circ}$ video\\ saliency prediction \cite{9072511} \\ - $360^{\circ}$ video summarization \cite{kontostathis2024summarization} (used to\\ evaluate $360^{\circ}$ video saliency prediction) \end{tabular}   &   Public    \\  \hline
\begin{tabular}[c]{@{}l@{}}$360^{\circ}$ video highlight \\ detection \cite{Yu2019DeepRanking}\end{tabular}                 & \begin{tabular}[c]{@{}l@{}}$2406$ NFOV and\\ $115$ $360^{\circ}$ videos\end{tabular}         & \begin{tabular}[c]{@{}l@{}}Wedding and \\ music videos\end{tabular}                                                                                                                              & \begin{tabular}[c]{@{}l@{}}$15\%$ most salient NFOV \\ video clips for $25$ $360^{\circ}$ \\videos from each category\end{tabular}                            & $360^{\circ}$ video highlight detection \cite{Yu2019DeepRanking}                                                                                       & Private                 \\ \hline
\begin{tabular}[c]{@{}l@{}}$360^{\circ}$ video story-based\\ summarization \cite{Lee_2018_CVPR}\end{tabular} & \begin{tabular}[c]{@{}l@{}}$285$ $360^{\circ}$ videos and\\ $9138$ photostreams\end{tabular} & \begin{tabular}[c]{@{}l@{}}Wedding, Parade,\\ Rowing, Scuba \\diving, Air ballooning\end{tabular}                                                       & \begin{tabular}[c]{@{}l@{}}Summaries for $50$ $360^{\circ}$\\ videos ($10$ videos per topic)\end{tabular}                                                  & \begin{tabular}[c]{@{}l@{}}Story-based $360^{\circ}$ video summarization \cite{Lee_2018_CVPR}\\ (used to evaluate temporal summarization)\end{tabular}      & Private                 \\ \hline
\end{tabular}}
\label{tab:datasets}
\end{table}

\section{The 360-VSumm Dataset}
\label{sec:dataset}
Our dataset was formulated based on the VR-EyeTracking dataset \cite{Xu_2018_CVPR}, which comprises of $208$ dynamic high-definition $360^{\circ}$ videos from YouTube. These videos were captured using both static and moving cameras and have a duration that spans from $20$ to $60$ seconds. Moreover, as mentioned in Section \ref{sec:literature}, they exhibit a diverse range of content, including indoor and outdoor scenes, underwater activities, sports games, short films, and more. Fixation maps were produced with the help of $45$ participants that utilized an HTC VIVE as their Head-Mounted Display (HMD) to immerse themselves in the $360^{\circ}$ videos, and using the Unity game engine to record the participants' head and gaze directions during their viewing experience. The fixation points were subsequently processed by convolving with a Gaussian filter with a standard deviation ($s=9.35$ degrees), as described in \cite{Atsal_2020}. This process resulted in a set of ground-truth saliency maps for the ERP frames of each $360^{\circ}$ video of the VR-EyeTracking dataset, that were used for constructing 360-VSumm.

\begin{figure}[t]
    \centering
    \includegraphics[width=0.8\textwidth]{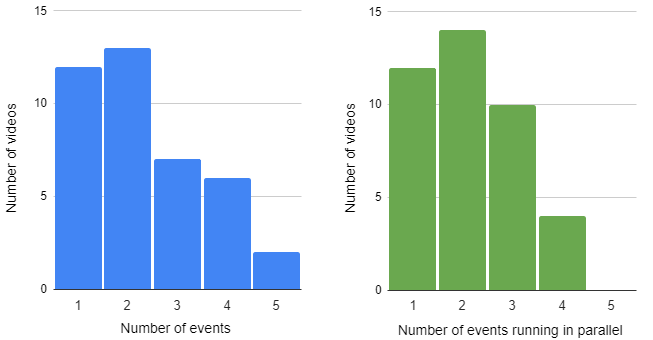}
    \caption{Histograms with the number of events per video (left side) and the number of events running in parallel per video (right side).}
    \label{fig:events_in_parallel}
\end{figure}

More specifically, we used the ERP frames and their associated ground-truth saliency maps to perform salient event detection and produce a 2D-video containing the detected events, following the algorithm described in \cite{kontostathis2024summarization}. This algorithm: i) identifies the salient regions in each frame by clustering salient points according to their intensity and distance, ii) defines spatial-temporally-correlated 2D sub-volumes by grouping together spatially related regions across a sequence of frames, iii) mitigates abrupt changes in the visual content of sub-volumes by adding possibly missing frames, iv) extracts the FOV for the salient regions of each ERP frame of the finally-formed sub-volume, v) forms the 2D-video fragments by cropping a spatial window around the salient regions of each ERP frame of the finally-formed sub-volumes, with an aspect ratio equal to $4:3$, and vi) produces the 2D-video by stitching the extracted 2D-video fragments in chronological order. In addition, this algorithm computes a saliency score for each frame of the produced 2D-video, by averaging the saliency of the identified regions within that frame. After performing the aforementioned process on the entire set of $360^{\circ}$ videos of the VR-EyeTracking dataset, we selected a subset of the created 2D-videos for further annotation in terms of video summarization, based on the following criteria: their visual content should be dynamic and diverse (in order to avoid videos that could be summarized using trivial approaches, such as random fragment sampling), and their duration should by longer than $1$ minute (in order to filter-out very short videos that are of limited interest in terms of summarization). Through this process, we ended up with $40$ 2D-videos with a duration ranging from $1$ to $4$ minutes. Most videos present more than one events (see the histogram on the left side of Fig. \ref{fig:events_in_parallel}), which typically overlap in time or run in parallel in the $360^{\circ}$ videos (see the histogram on the right side of Fig. \ref{fig:events_in_parallel}).

\begin{figure}[t]
    \centering
    \includegraphics[width=1\textwidth]{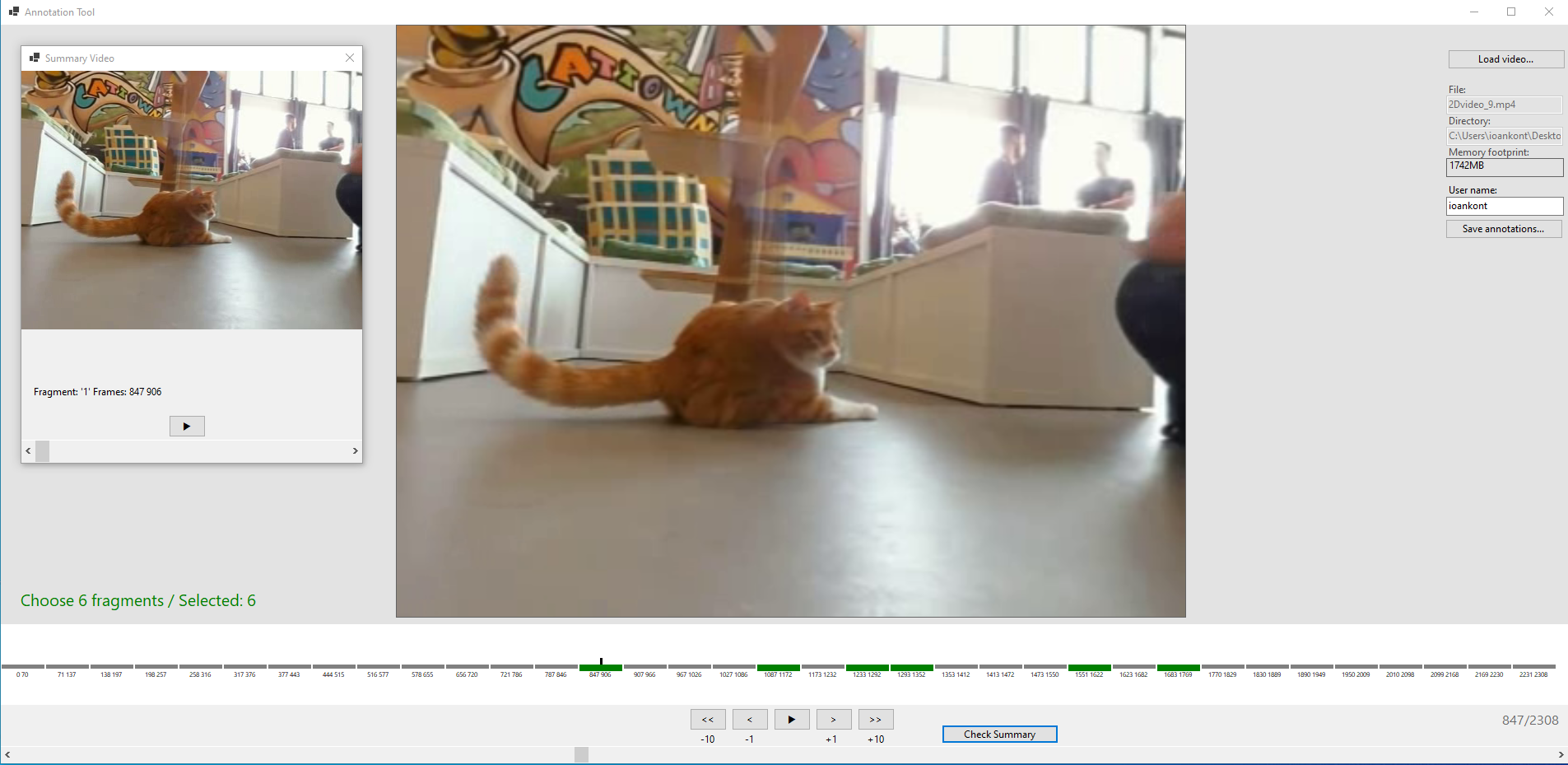}
    \caption{The graphical interface of the developed annotation tool.}
    \label{fig:annotation_tool}
\end{figure}

The selected videos were annotated by $15$ annotators ($13$ males $2$ females with ages between $24-44$), that were asked to pick the most important/interesting or informative parts of each video in order to form a summary with a duration that is approx. equal to the $15\%$ of the 2D-video's length. As stated above, each 2D-video is composed of a set of fragments presenting the detected salient events in the corresponding $360^{\circ}$ video. These fragments were further segmented into non-overlapping and roughly equal in length ($2$ sec.) sub-fragments, in order to allow the annotators to select specific parts of a given fragment for inclusion in the video summary rather than selecting the whole fragment. So, each 2D-video was segmented into M sub-fragments and the task of the annotators was to select N of these sub-fragments for inclusion in the summary, where N equals to $15\%$ of M. To facilitate the annotation process, we developed a tool with a user-friendly graphical interface that is depicted in Fig. \ref{fig:annotation_tool}. To start the annotation process, the participants have to load a video from the collection by clicking on the corresponding button on the upper right part of the interface. Subsequently, they see the first frame of the selected video in the video player of the tool and the sub-fragments of the video in the white-coloured area between the video player and the navigation buttons. Each dash in this area corresponds to a different sub-fragment and it is clickable to allow the selection of the relevant part of the video. Moreover, through the user interface the annotators are notified about the number of sub-fragments that should be selected in order to form a summary according to the targeted time budget. When the necessary number of sub-fragments is reached, the annotators can immediately check the formed summary by clicking on the relevant button. This action opens a new smaller video player (which appears in the left side of the main video player, in Fig. \ref{fig:annotation_tool}) that plays only the chosen parts of the video. If the annotator needs to make changes in the summary, s/he can replace one or more of the selected sub-fragments, by clicking once more on the relevant dashes (to un-select them) and click on other dashes (to select them). Through this process, the annotators can update the summary as many times as they wish, and can re-check it by clicking on the ``Check Summary'' button. After concluding to an optimal video summary, they have to click on the ``Save annotations...'' button in order to store their preferences for the annotated video. This process produces a txt file per video, containing information about the starting and ending frame of each selected sub-fragment of the video. As a note, the review of the formed summary is possible only when the correct number of sub-fragments has been selected. If the selected sub-fragments are less/more that the suggested ones, the relevant text in the user interface is coloured gray/red, while after clicking on the ``Check Summary'' button, the annotator is notified that ``the number of selected fragments is lower/higher than the suggested one''. Moreover, the interface is equipped with the standard functionalities for navigation in the video, i.e., buttons for play/stop, frame-by-frame display (``>'' and ``<''), and $10$ frames skip (``<<''  and ``>>''), while the annotators can navigate also via the horizontal scroll bar at the bottom of the interface.

The created 360-VSumm dataset contains: 
\begin{itemize}
    \item $40$ 2D-videos that were created as described in the second paragraph of this section.
    \item $15$ ground-truth human-generated summaries per video (binary vectors indicating which frames have been selected for inclusion in the summary and which not).
    \item A mean ground-truth summary that can be used for supervised training (obtained after averaging the $15$ ground-truth summaries at the frame-level).
    \item Data about the fragments of these videos, as defined by the applied salient event detection and 2D-video production algorithm.
    \item Data about the sub-fragments of these videos, that were used during the annotation process.
    \item Data about the saliency of each frame of these videos.
\end{itemize}
The 360-VSumm dataset is publicly-available at: \url{https://github.com/IDT-ITI/360-VSumm}

\section{Experiments}

Building on the created 360-VSumm dataset, we investigated the following research questions:
\begin{itemize}
    \item Can we use pre-trained models of state-of-the-art methods for conventional video summarization, to produce summaries for $360^{\circ}$ videos?
    \item Are there any observed performance gains after re-training these models using data from $360^{\circ}$ videos?
    \item Does it help to utilize information about the saliency of the visual content under summarization?
\end{itemize}

To answer these questions, we considered two state-of-the-art methods for (traditional) video summarization from the literature, that learn how to estimate the importance of video frames using attention mechanisms. Based on their estimates, the video summary is formed by computing fragment-level importance (by averaging the importance of the frames that lie within each fragment) and selecting the top-scoring fragments so as to create a summary that lasts equally to the $15\%$ of the video duration. The first method is called PGL-SUM \cite{9666088} and learns the task in a supervised manner using ground-truth annotations. PGL-SUM combines global and local multi-head attention mechanisms to discover different modelings of the frames' dependence at different levels of granularity. Moreover, the utilized attention mechanisms integrate a component that encodes the temporal position of video frames, as this is of major importance when producing a video summary. The second method is called CA-SUM \cite{10.1145/3512527.3531404} and learns the task in a fully-unsupervised way based on a simple loss function that relates to the length of the generated video summary. CA-SUM integrates a concentrated attention mechanism that is able to focus on non-overlapping blocks in the main diagonal of the attention matrix, and to incorporate knowledge about the uniqueness and diversity of the associated frames of the video. 

\subsection{Implementation Details}

Following the typical approach in the video summarization literature \cite{9594911}, videos were downsampled to 2 fps and the sampled frames are represented using the output of the pool5 layer of GoogleNet \cite{7298594} trained on ImageNet. With respect to PGL-SUM, we initialized the network's weights using the Xavier uniform initialization approach with gain = $\sqrt{2}$ and biases = $0.1$, we set the learning rate, dropout rate and L2 regularization factor equal to $5\cdot10^{-5}$, $0.5$ and $10^{-5}$ (as in \cite{9666088}), and we trained the model for $600$ epochs in a full-batch mode using the Adam optimizer. The utilized pre-trained models on the SumMe \cite{10.1007/978-3-319-10584-0_33} and TVSum \cite{7299154} benchmarking datasets for video summarization, were obtained from \url{https://zenodo.org/records/5635735} (we picked the ones under ``table3 models/dataset/split4'' that exhibited the best performance. With regards to CA-SUM, we used the same network initialization approach with PGL-SUM, we set the learning rate and the L2 regularization factor equal to $5\cdot10^{-4}$ and $10^{-5}$, respectively (similarly to \cite{10.1145/3512527.3531404}), and we trained the model for 600 epochs in a full-batch mode using the Adam optimizer. The utilized pre-trained models on the SumMe and TVSum datasets were downloaded from \url{https://zenodo.org/records/6562992} (we picked the models under ``SumMe/split3'' and ``TVSum/split4'', as these were the top-performing ones).

In terms of evaluation, we estimate the similarity between a machine-generated and a user-defined summary by computing their overlap using the F-Score (as percentage), following the typical evaluation protocol in the literature \cite{9594911}. So, for a given test video we compare the generated summary with the available user summaries for this video and compute an F-Score for each pair of compared summaries. Instead of computing the average value across the computed F-Scores, we then keep the maximum value as the final F-Score for this video, based on the intuition that a video summarization method should be able to produce a summary that resembles (at least) one of the user summaries. The computed F-Scores for all test videos are averaged and form the method's performance on the test set. Finally, following the paradigm in the relevant literature \cite{9594911}, we split the dataset into five different splits to perform 5-fold cross validation. In each split, $80\%$ of the videos are used for training and the remaining $20\%$ for testing, while there are no overlaps between the tests sets. In the following, we report the average performance of each of the considered methods, over these runs.

All experiments were carried out on an NVIDIA RTX 3090 GPU. The code for reproducing the reported results is publicly-available at: \url{https://github.com/IDT-ITI/360-VSumm}

\subsection{Quantitative Results}

To examine the extent to which pre-trained models of PGL-SUM and CA-SUM on the most commonly used datasets in the video summarization literature (SumMe \cite{10.1007/978-3-319-10584-0_33} and TVSum \cite{7299154}) can be utilized to produce summaries for $360^{\circ}$ videos, we measured their performance on the videos of the created 360-VSumm dataset and compared it with the performance of a random summarizer. The performance of a random summarizer on a given video was measured as proposed in \cite{10.1145/3394171.3413632}. In particular, we initially assigned randomly-created importance scores to the video frames based on a uniform distribution of probabilities. Following, we computed fragment-level scores based on the predefined segmentation of the video, and we formed the video summary by selecting the $15\%$ top-scoring fragments. Random summarization was performed $100$ times and we report the average score over these runs. The values in Table \ref{tab:pretrained_models} show that models of PGL-SUM and CA-SUM that have been trained for conventional 2D-video summarization, cannot be used for summarizing $360^{\circ}$ videos as they exhibit random-level performance on the videos of the created 360-VSumm dataset. Such an experimental setting is taken into account in several works on video summarization (e.g., \cite{LIANG2022108840,10.1007/s11042-022-12901-4,ZHU2022108312,ZANG202326,ZHANG2024123568}) and called ``transfer setting''. The reported results in these works show that this setting is more challenging than the canonical setting where a model is trained and evaluated on the same dataset, as a drop in performance is usually observed for the other (unseen during training) dataset; nevertheless, the recorded performance is clearly higher than the performance of a random summarizer. These findings shows that the summarization of $360^{\circ}$ videos is a different and more challenging task, and needs methods that are better tailored to the visual characteristics of such videos. 

\begin{table}[t]
\centering
\caption{The performance (F-Score (\%)) of a random summarizer and two pre-trained models of PGL-SUM and CA-SUM, on the 360-VSumm dataset.}
\begin{tabular}{|l|c|c|}
\hline
Method                   & \begin{tabular}[c]{@{}c@{}}Pre-training\\ dataset\end{tabular} & \begin{tabular}[c]{@{}c@{}}Performance\\ (F-Score (\%))\end{tabular} \\ \hline
\multirow{2}{*}{PGL-SUM} & SumMe                                                          & 33.7                                                                 \\
                         & TVSum                                                          & 33.2                                                                 \\ \hline
\multirow{2}{*}{CA-SUM}  & SumMe                                                          & 36.5                                                                 \\
                         & TVSum                                                          & 35.4                                                                 \\ \hline
Random                   & -                                                              & 35.2                                                                 \\ \hline
\end{tabular}
\label{tab:pretrained_models}
\end{table}

\begin{table}[t]
\centering
\caption{The performance (F-Score (\%)) of PGL-SUM after being trained on the 360-VSumm dataset, for different configurations with respect to the number of local attention mechanisms (upper part) and the number of global and local attention heads (lower part).}
\begin{tabular}{|lccc|}
\hline
\multicolumn{4}{|l|}{}                                                                                                                                           \\ [-0.7em]
\multicolumn{4}{|l|}{Study on local attention mechanisms}                                                                                                        \\ [-0.7em]
\multicolumn{4}{|l|}{}                                                                                                                                           \\ \hline
\multicolumn{1}{|l|}{\begin{tabular}[c]{@{}l@{}}\# of local \\ attention\end{tabular}}   & \multicolumn{1}{c|}{2}    & \multicolumn{1}{c|}{4}    & 8             \\ \hline
\multicolumn{1}{|l|}{\begin{tabular}[c]{@{}l@{}}Performance\\ (F-Score (\%))\end{tabular}} & \multicolumn{1}{c|}{46.0} & \multicolumn{1}{c|}{46.0} & \textbf{46.3} \\ \hline
\multicolumn{4}{|l|}{}                                                                                                                                           \\ [-0.7em]
\multicolumn{4}{|l|}{Study on attention heads}                                                                                                                   \\ [-0.7em]
\multicolumn{4}{|l|}{}                                                                                                                                           \\ \hline
\multicolumn{1}{|l|}{\diagbox{Global}{Local}}                                        & \multicolumn{1}{c|}{2}    & \multicolumn{1}{c|}{4}    & 8             \\ \hline
\multicolumn{1}{|l|}{2}                                                                  & \multicolumn{1}{c|}{46.6} & \multicolumn{1}{c|}{46.5} & 47.3          \\ \hline
\multicolumn{1}{|l|}{4}                                                                  & \multicolumn{1}{c|}{47.3} & \multicolumn{1}{c|}{46.3} & 46.4          \\ \hline
\multicolumn{1}{|l|}{8}                                                                  & \multicolumn{1}{c|}{46.0} & \multicolumn{1}{c|}{46.3} & 45.8          \\ \hline
\multicolumn{1}{|l|}{16}                                                                 & \multicolumn{1}{c|}{46.3} & \multicolumn{1}{c|}{45.9} & \textbf{47.9} \\ \hline
\end{tabular}
\label{tab:attention}
\end{table}

Subsequently, to quantify possible gains in the summarization performance after training these models with data from $360^{\circ}$ videos, we re-trained and evaluated models of PGL-SUM and CA-SUM using the formed data splits for 360-VSumm. Instead of taking into account only the default values for the hyper-parameters of these models (as reported in \cite{10.1145/3512527.3531404,9666088}), we experimented with various options aiming to find the top-performing configurations. For \textbf{PGL-SUM}, we started by trying to find the optimal number of local attention mechanism and continued by investigating different choices about the number of attention heads (similarly to the methodology in \cite{10.1145/3512527.3531404}). The results in the upper part of Table \ref{tab:attention} indicate that a higher (and maximum supported) number of local attention mechanisms than the one in the original PGL-SUM model (equal to four) improves the model's effectiveness on the 360-VSumm dataset, as it leads to higher summarization performance. Furthermore, the examined combinations about the number of attention heads demonstrate that the use of the maximum allowed heads for the global and the eight local attention mechanisms, results in a further performance improvement by $1.6\%$ (see the lower part of Table \ref{tab:attention}). For \textbf{CA-SUM}, we initially ran experiments for the different values of the regularization factor sigma that are taken into account by the original method, and then we focused on the block size of the concentrated attention mechanism. The reported results in the upper part of Table \ref{tab:reg_block} show that the best performance for the default block size of CA-SUM (which was set equal to $60$ based on the restrictions of the SumMe and TVSum datasets) is obtained after setting the regularization factor equal to $0.7$. Subsequent evaluations using this regularization factor show that a larger block size (equal to $70$) favors the learning process, as it further advances the summarization performance by $0.6\%$. Based on the findings reported above, we argue that using the created 360-VSumm dataset for training models that have been designed for conventional video summarization, can lead to noticeable performance gains that clearly distinguish these models from baseline (random) approaches. In addition, a performance comparison between the PGL-SUM and CA-SUM demonstrates that using a supervision signal for training is beneficial, as it leads to improved summarization performance.

\begin{table}[t]
\centering
\caption{The performance (F-Score (\%)) of CA-SUM after being trained on the 360-VSumm dataset, for different values of the regularization factor sigma (upper part) and the block size of the concentrated attention mechanism (lower part).}
\begin{tabular}{|lccccc|}
\hline
\multicolumn{6}{|l|}{}                                                                                                                                                                                                     \\ [-0.7em]
\multicolumn{6}{|l|}{Study about the regularization factor sigma}                                                                                                                                                          \\ [-0.7em]
\multicolumn{6}{|l|}{}                                                                                                                                                                                                     \\ \hline
\multicolumn{1}{|l|}{Regular. factor}                                                      & \multicolumn{1}{c|}{0.5}  & \multicolumn{1}{c|}{0.6}  & \multicolumn{1}{c|}{0.7}           & \multicolumn{1}{c|}{0.8}  & 0.9  \\ \hline
\multicolumn{1}{|l|}{\begin{tabular}[c]{@{}l@{}}Performance\\ (F-Score (\%))\end{tabular}} & \multicolumn{1}{c|}{44.2} & \multicolumn{1}{c|}{43.5} & \multicolumn{1}{c|}{\textbf{44.7}} & \multicolumn{1}{c|}{44.4} & 43.9 \\ \hline
\multicolumn{6}{|l|}{}                                                                                                                                                                                                     \\ [-0.7em]
\multicolumn{6}{|l|}{Study about the block size of the attention}                                                                                                                                                          \\ [-0.7em]
\multicolumn{6}{|l|}{}                                                                                                                                                                                                     \\ \hline
\multicolumn{1}{|l|}{Block size}                                                           & \multicolumn{1}{c|}{50}   & \multicolumn{1}{c|}{60}   & \multicolumn{1}{c|}{70}            & \multicolumn{1}{c|}{80}   & 90   \\ \hline
\multicolumn{1}{|l|}{\begin{tabular}[c]{@{}l@{}}Performance\\ (F-Score (\%))\end{tabular}} & \multicolumn{1}{c|}{44.2} & \multicolumn{1}{c|}{44.7} & \multicolumn{1}{c|}{\textbf{45.3}} & \multicolumn{1}{c|}{44.6} & 43.7 \\ \hline
\end{tabular}
\label{tab:reg_block}
\end{table}

Finally, we investigated the influence of using auxiliary information about the saliency of the visual content, during the training of the video summarization models. For this, we evaluated the performance of variants of PGL-SUM and CA-SUM that use the computed saliency scores for the frames of the produced 2D-videos (as described in the second paragraph of Section \ref{sec:dataset}) to weight the extracted representations of the visual content of these frames. Thus, these variants incorporate information about both the visual content and the saliency of each frame. The outcomes of the conducted evaluations, reported in Table \ref{tab:saliency}, indicate the positive impact of using the frames' saliency when summarizing $360^{\circ}$ videos, since it leads to increased summarization performance for both PGL-SUM and CA-SUM methods.

\begin{table}[t]
\centering
\caption{The performance (F-Score (\%)) of the top-performing models of PGL-SUM and CA-SUM, after being trained using also information about the visual saliency of the video frames.}
\begin{tabular}{|lc|}
\hline
\multicolumn{1}{|l|}{Method}                                                      & \begin{tabular}[c]{@{}c@{}}Performance\\ (F-Score (\%))\end{tabular} \\ \hline
\multicolumn{1}{|l|}{\begin{tabular}[c]{@{}l@{}}Random\\ Summarizer\end{tabular}} & 35.2                                                                 \\ \hline
                                                                                  & \multicolumn{1}{l|}{}                                                \\ [-0.8em] \hline
\multicolumn{1}{|l|}{PGL-SUM}                                                     & 47.9                                                                 \\
\multicolumn{1}{|l|}{PGL-SUM-sal}                                                 & \textbf{48.2}                                                                 \\ \hline
                                                                                  & \multicolumn{1}{l|}{}                                                \\ [-0.8em] \hline
\multicolumn{1}{|l|}{CA-SUM}                                                      & 45.3                                                                 \\
\multicolumn{1}{|l|}{CA-SUM-sal}                                                  & \textbf{46.6}                                                                 \\ \hline
\end{tabular}
\label{tab:saliency}
\end{table}

\subsection{Qualitative Results}

\begin{figure}[t]
    \centering
    \includegraphics[width=\textwidth]{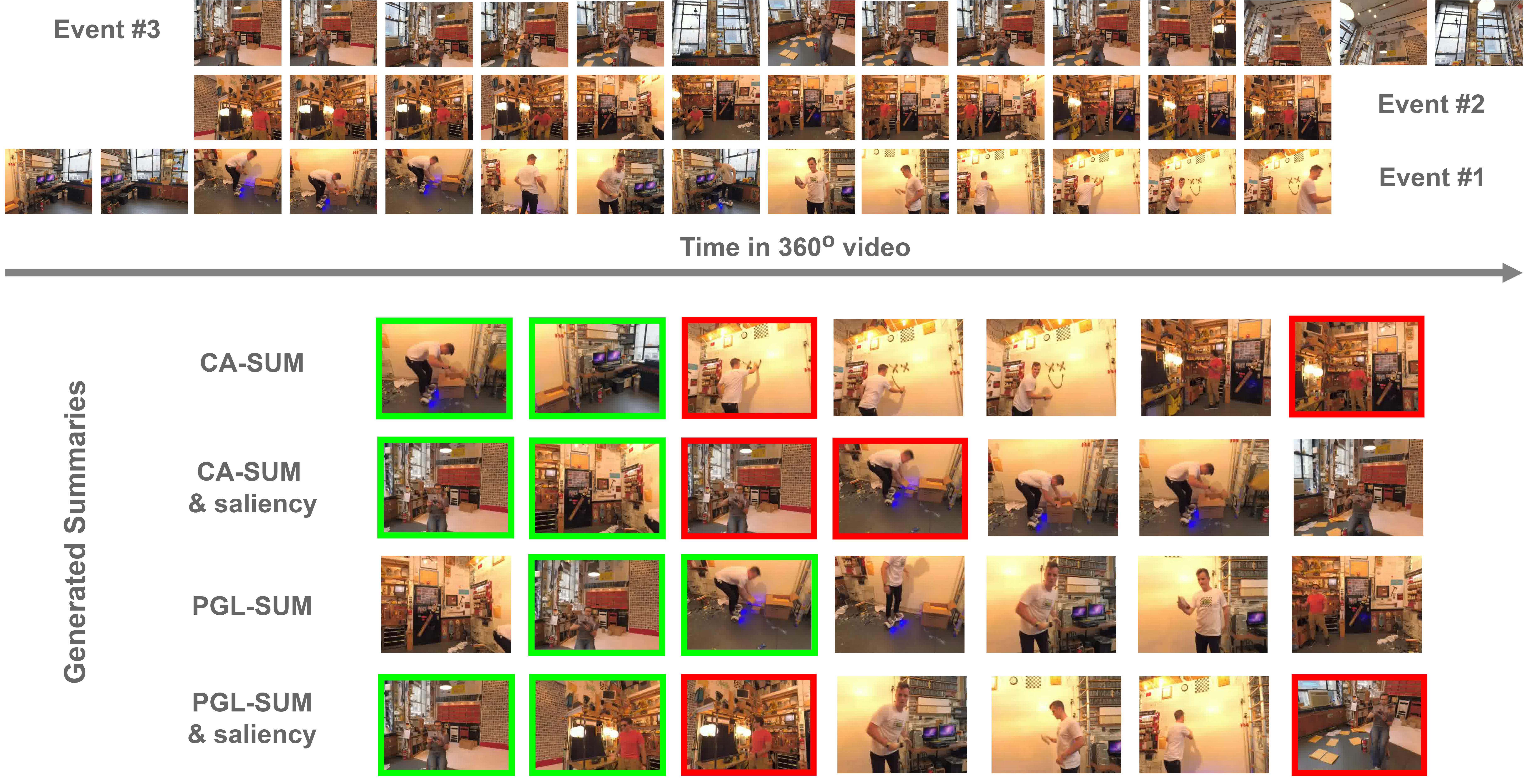}
    \caption{A frame-based overview of the presented events in the video (top part), and the produced summaries by the best-performing models of CA-SUM, PGL-SUM and their saliency-aware variants (bottom part).}
    \label{fig:example_1}
\end{figure}

\begin{figure}[t]
    \centering
    \includegraphics[width=\textwidth]{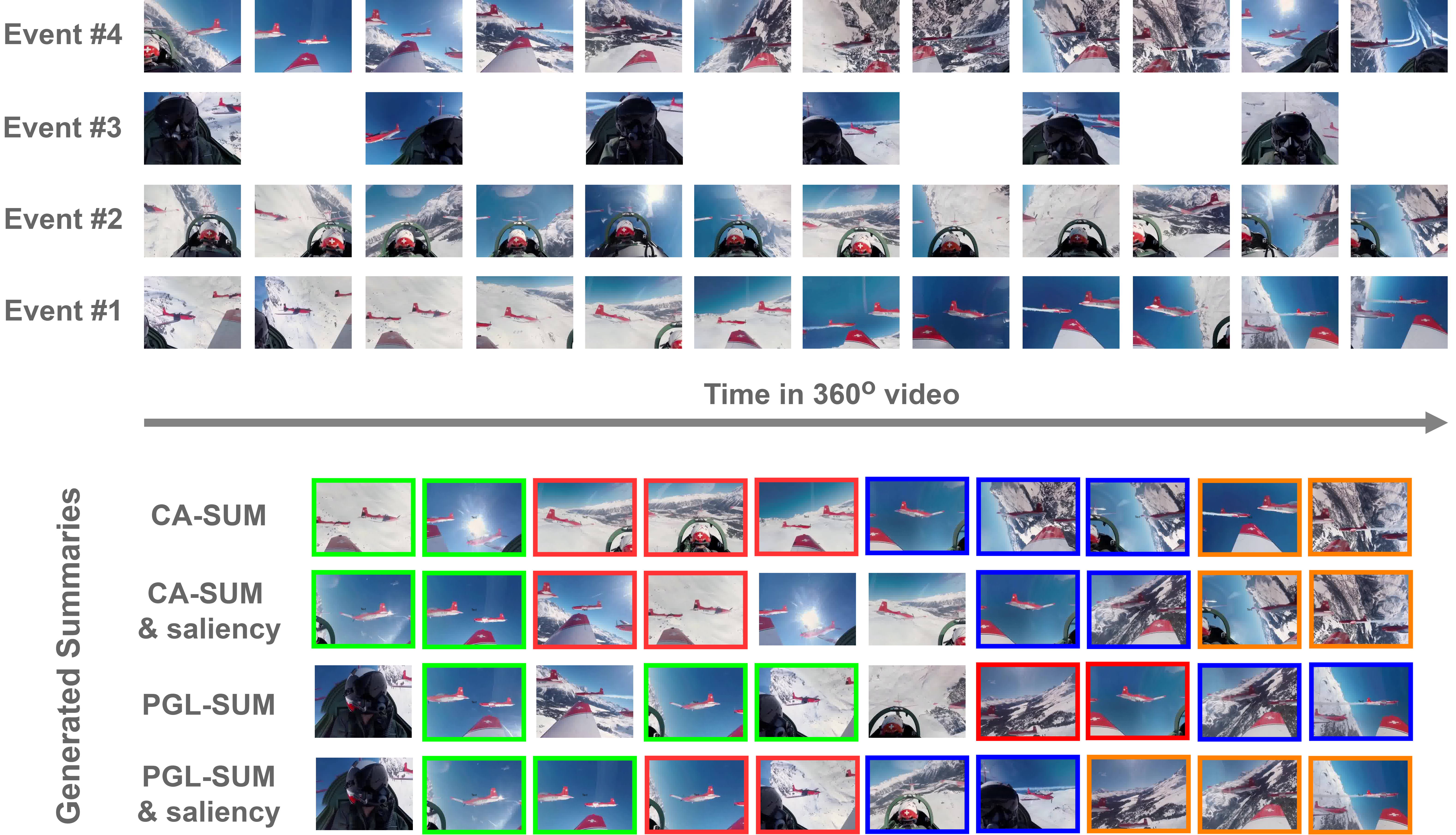}
    \caption{A frame-based overview of the presented events in the video (top part), and the produced summaries by the best-performing models of CA-SUM, PGL-SUM and their saliency-aware variants (bottom part).}
    \label{fig:example_2}
\end{figure}

In addition to the above reported quantitative evaluations, we conducted a qualitative analysis using two indicative videos of the created 360-VSumm dataset and examining the produced video summaries by the best-performing models of CA-SUM, PGL-SUM and their saliency-aware variants. Both videos present multiple events that either overlap in time or run in parallel in the original $360^{\circ}$ video. We argue that the summarization of such videos is a more challenging task compared to the summarization of a video showing a single event, as an optimal video summary should contain information about all the events in the full-length video. Our analysis was based on the examples depicted in Fig. \ref{fig:example_1} and \ref{fig:example_2}. In both of these figures, the top part provides a frame-based overview of the presented events in the videos after selecting some representative keyframes (for space reasons). The bottom part contains the created video summaries by the aforementioned models, that are presented using a number of representative keyframes (for the sake of space, we excluded visually-similar frames from these illustrations). 

Concerning the first example, as shown in Fig. \ref{fig:example_1}, the created summary by CA-SUM focuses mainly on the first event of the video, contains a small part of the second event, and is missing information about the third event. The produced summary after taking into account the frames' saliency (CA-SUM-sal) includes mainly introductory parts from the first and third event, and lacks information about the second event. The PGL-SUM method appears to be more effective, as the created summary comprises of segments that correspond to all three events of the video. However, it mainly focuses on the first event and fails to convey information about the main activity in the remaining two. Incorporating information also about the frames' saliency (PGL-SUM-sal) leads to a more complete video summary, that covers all the different events in a more balanced and descriptive way. With respect to the second example, as illustrated in Fig. \ref{fig:example_2}, CA-SUM forms a video summary which presents mainly the activity at the left and right side of the airplane that carries the video camera (first and fourth event), contains some views from the back of the pilot (second event), and completely misses the front view of the pilot (third event). Similar remarks can be made for CA-SUM-sal, with the observation that the second event is less pronounced in the produced summary. With respect to the remaining methods (PGL-SUM, PGL-SUM-sal), the created summaries contain parts from all the different events, thus proving a complete overview of the video. The findings reported above are aligned with the outcomes of our quantitative evaluations, as they indicate the advanced performance of PGL-SUM compared to CA-SUM and demonstrate the positive impact of extracting and using information about the frames' saliency, when summarizing  $360^{\circ}$ videos. 

\section{Conclusions and Next Steps}

In this paper we presented the 360-VSumm dataset that can be used for training and evaluating methods for $360^{\circ}$ video summarization. Based on the created dataset, we trained two state-of-the-art methods for conventional 2D-video summarization and assessed their performance to form a baseline for future comparisons. Moreover, we took into account two saliency-aware variants of these methods and documented the positive impact of incorporating information about the frames' saliency during the summarization process. Finally, we presented the developed interactive tool for annotation purposes, that can be used to facilitate similar annotation activities. In the future, we will extract additional information about the frames of the produced 2D-videos (such as their spatial positioning in the $360^{\circ}$ video) and use it as extra auxiliary data for training $360^{\circ}$ video summarization methods.

\section*{Acknowledgments}
This work was supported by the EU Horizon Europe and Horizon 2020 programmes under grant agreements 101070109 TransMIXR and 951911 AI4Media, respectively.


\begin{thebibliography}{10}

\bibitem{kontostathis2024summarization}
Ioannis Kontostathis, Evlampios Apostolidis, and Vasileios Mezaris.
\newblock An integrated system for spatio-temporal summarization of 360-degrees videos.
\newblock In Stevan Rudinac, Alan Hanjalic, Cynthia Liem, Marcel Worring, Bj{\"o}rn~Þ{\'o}r J{\'o}nsson, Bei Liu, and Yoko Yamakata, editors, {\em MultiMedia Modeling}, pages 202--215, Cham, 2024. Springer Nature Switzerland.

\bibitem{su2016activity}
Yu-Chuan Su, Dinesh Jayaraman, and Kristen Grauman.
\newblock Pano2vid: Automatic cinematography for watching 360 videos.
\newblock In {\em Proc. of the Asian Conference on Computer Vision (ACCV)}, 2016.

\bibitem{Su_2017_CVPR}
Yu-Chuan Su and Kristen Grauman.
\newblock Making 360deg video watchable in 2d: Learning videography for click free viewing.
\newblock In {\em Proc. of the IEEE Conf. on Computer Vision and Pattern Recognition (CVPR)}, 2017.

\bibitem{Hu_2017_CVPR}
Hou-Ning Hu, Yen-Chen Lin, Ming-Yu Liu, Hsien-Tzu Cheng, Yung-Ju Chang, and Min Sun.
\newblock Deep 360 pilot: Learning a deep agent for piloting through 360deg sports videos.
\newblock In {\em Proc. of the IEEE Conf. on Computer Vision and Pattern Recognition (CVPR)}, 2017.

\bibitem{9072511}
Minglang Qiao, Mai Xu, Zulin Wang, and Ali Borji.
\newblock Viewport-dependent saliency prediction in 360° video.
\newblock {\em IEEE Transactions on Multimedia}, 23:748--760, 2021.

\bibitem{10.1145/3306346.3323046}
Kyoungkook Kang and Sunghyun Cho.
\newblock Interactive and automatic navigation for 360° video playback.
\newblock {\em ACM Trans. Graph.}, 38(4), 2019.

\bibitem{9284734}
Miao Wang, Yi-Jun Li, Wen-Xuan Zhang, Christian Richardt, and Shi-Min Hu.
\newblock Transitioning360: Content-aware nfov virtual camera paths for 360° video playback.
\newblock In {\em 2020 IEEE International Symposium on Mixed and Augmented Reality (ISMAR)}, pages 185--194, 2020.

\bibitem{Yu2019DeepRanking}
Youngjae Yu, Sangho Lee, Joonil Na, Jaeyun Kang, and Gunhee Kim.
\newblock {A Deep Ranking Model for Spatio-Temporal Highlight Detection From a 360 Video}.
\newblock In {\em Proc. of the 2018 {AAAI} Conf. on Artificial Intelligence}, 2018.

\bibitem{Lee_2018_CVPR}
Sangho Lee, Jinyoung Sung, Youngjae Yu, and Gunhee Kim.
\newblock A memory network approach for story-based temporal summarization of 360° videos.
\newblock In {\em Proc. of the IEEE Conf. on Computer Vision and Pattern Recognition (CVPR)}, June 2018.

\bibitem{8418756}
Mai Xu, Yuhang Song, Jianyi Wang, Minglang Qiao, Liangyu Huo, and Zulin Wang.
\newblock Predicting head movement in panoramic video: A deep reinforcement learning approach.
\newblock {\em IEEE Transactions on Pattern Analysis and Machine Intelligence}, 41(11):2693--2708, 2019.

\bibitem{Xu_2018_CVPR}
Yanyu Xu, Yanbing Dong, Junru Wu, Zhengzhong Sun, Zhiru Shi, Jingyi Yu, and Shenghua Gao.
\newblock Gaze prediction in dynamic 360° immersive videos.
\newblock In {\em 2018 IEEE/CVF Conference on Computer Vision and Pattern Recognition}, pages 5333--5342, 2018.

\bibitem{Atsal_2020}
Yasser Dahou, Marouane Tliba, Kevin McGuinness, and Noel O'Connor.
\newblock Atsal: An attention based architecture for saliency prediction in 360 videos.
\newblock In Alberto Del~Bimbo, Rita Cucchiara, Stan Sclaroff, Giovanni~Maria Farinella, Tao Mei, Marco Bertini, Hugo~Jair Escalante, and Roberto Vezzani, editors, {\em Pattern Recognition. ICPR International Workshops and Challenges}, pages 305--320, Cham, 2021. Springer International Publishing.

\bibitem{Sstsal_2022_CG}
Edurne Bernal-Berdun, Daniel Martin, Diego Gutierrez, and Belen Masia.
\newblock {SST-Sal: A} spherical spatio-temporal approach for saliency prediction in 360 videos.
\newblock {\em Computers \& Graphics}, 106:200--209, 2022.

\bibitem{10.1145/3512527.3531404}
Evlampios Apostolidis, Georgios Balaouras, Vasileios Mezaris, and Ioannis Patras.
\newblock Summarizing videos using concentrated attention and considering the uniqueness and diversity of the video frames.
\newblock In {\em Proceedings of the 2022 International Conference on Multimedia Retrieval}, ICMR '22, page 407–415, New York, NY, USA, 2022. Association for Computing Machinery.

\bibitem{9666088}
Evlampios Apostolidis, Georgios Balaouras, Vasileios Mezaris, and Ioannis Patras.
\newblock Combining global and local attention with positional encoding for video summarization.
\newblock In {\em 2021 IEEE International Symposium on Multimedia (ISM)}, pages 226--234, 2021.

\bibitem{9594911}
Evlampios Apostolidis, Eleni Adamantidou, Alexandros~I. Metsai, Vasileios Mezaris, and Ioannis Patras.
\newblock Video summarization using deep neural networks: A survey.
\newblock {\em Proceedings of the IEEE}, 109(11):1838--1863, 2021.

\bibitem{7298594}
Christian Szegedy, Wei Liu, Yangqing Jia, Pierre Sermanet, Scott Reed, Dragomir Anguelov, Dumitru Erhan, Vincent Vanhoucke, and Andrew Rabinovich.
\newblock Going deeper with convolutions.
\newblock In {\em 2015 IEEE Conference on Computer Vision and Pattern Recognition (CVPR)}, pages 1--9, 2015.

\bibitem{10.1007/978-3-319-10584-0_33}
Michael Gygli, Helmut Grabner, Hayko Riemenschneider, and Luc Van~Gool.
\newblock {Creating Summaries from User Videos}.
\newblock In David Fleet, Tomas Pajdla, Bernt Schiele, and Tinne Tuytelaars, editors, {\em Europ. Conf. on Computer Vision (ECCV) 2014}, pages 505--520, Cham, 2014. Springer International Publishing.

\bibitem{7299154}
Yale Song, J.~Vallmitjana, A.~Stent, and A.~Jaimes.
\newblock {TVSum: Summarizing web videos using titles}.
\newblock In {\em 2015 IEEE/CVF Conf. on Computer Vision and Pattern Recognition (CVPR)}, pages 5179--5187, June 2015.

\bibitem{10.1145/3394171.3413632}
Evlampios Apostolidis, Eleni Adamantidou, Alexandros~I. Metsai, Vasileios Mezaris, and Ioannis Patras.
\newblock {Performance over Random: A Robust Evaluation Protocol for Video Summarization Methods}.
\newblock In {\em Proc. of the 28th ACM Int. Conf. on Multimedia (MM '20)}, page 1056–1064, New York, NY, USA, 2020. ACM.

\bibitem{LIANG2022108840}
Guoqiang Liang, Yanbing Lv, Shucheng Li, Shizhou Zhang, and Yanning Zhang.
\newblock Video summarization with a convolutional attentive adversarial network.
\newblock {\em Pattern Recognition}, 131:108840, 2022.

\bibitem{10.1007/s11042-022-12901-4}
H.~Min, H.~Ruimin, W.~Zhongyuan, Xi. Zixiang, and Z.~Rui.
\newblock Spatiotemporal two-stream lstm network for unsupervised video summarization.
\newblock {\em Multimedia Tools and Applications}, 81:40489--40510, 2022.

\bibitem{ZHU2022108312}
Wencheng Zhu, Jiwen Lu, Yucheng Han, and Jie Zhou.
\newblock Learning multiscale hierarchical attention for video summarization.
\newblock {\em Pattern Recognition}, 122:108312, 2022.

\bibitem{ZANG202326}
Sha-Sha Zang, Hui Yu, Yan Song, and Ru~Zeng.
\newblock Unsupervised video summarization using deep non-local video summarization networks.
\newblock {\em Neurocomputing}, 519:26--35, 2023.

\bibitem{ZHANG2024123568}
Yunzuo Zhang, Yameng Liu, and Cunyu Wu.
\newblock Attention-guided multi-granularity fusion model for video summarization.
\newblock {\em Expert Systems with Applications}, 249:123568, 2024.

\end{thebibliography}

\end{document}